\documentclass{midl} % Include author names
\usepackage{ulem}
\usepackage{adjustbox}
\usepackage{mwe} % to get dummy images
\jmlryear{2023}
\jmlrworkshop{Full Paper -- MIDL 2023 submission}
\title[Attri-Net]{Inherently Interpretable Multi-Label Classification Using Class-Specific Counterfactuals}

\midlauthor{\Name{Susu Sun\nametag{$^{1}$}} \Email{susu.sun@uni-tuebingen.de}\\
\addr $^{1}$ Cluster of Excellence -- Machine Learning for Science, University of Tübingen, Germany \\
\Name{Stefano Woerner\nametag{$^{1}$}} \Email{stefano.woerner@uni-tuebingen.de}\\
\Name{Andreas Maier\nametag{$^{2}$}} \Email{andreas.maier@fau.de}\\
\addr $^{2}$ Pattern Recognition Lab, Friedrich-Alexander-Universität Erlangen-Nürnberg, Germany\\
\Name{Lisa M. Koch\nametag{$^{3, 4}$}} \Email{lisa.koch@uni-tuebingen.de}\\
\addr $^{3}$ Hertie Institute for Artificial Intelligence in Brain Health, University of T{\"u}bingen, Germany\\
\addr $^{4}$ Institute of Ophthalmic Research, University of T{\"u}bingen, Germany\\
\Name{Christian F. Baumgartner\nametag{$^{1}$}} \Email{christian.baumgartner@uni-tuebingen.de}\\
}

\DeclareMathOperator*{\E}{\mathbb{E}}

\begin{document}

\maketitle

\begin{abstract}
Interpretability is essential for machine learning algorithms in high-stakes application fields such as medical image analysis. However, high-performing black-box neural networks do not provide explanations for their predictions, which can lead to mistrust and suboptimal human-ML collaboration. Post-hoc explanation techniques, which are widely used in practice, have been shown to suffer from severe conceptual problems. Furthermore, as we show in this paper, current explanation techniques do not perform adequately in the multi-label scenario, in which multiple medical findings may co-occur in a single image. We propose Attri-Net\footnote{The code for Attri-Net is available at \url{https://github.com/ss-sun/Attri-Net}}, an inherently interpretable model for multi-label classification. Attri-Net is a powerful classifier that provides transparent, trustworthy, and human-understandable explanations. The model first generates class-specific attribution maps based on counterfactuals to identify which image regions correspond to certain medical findings. Then a simple logistic regression classifier is used to make predictions based solely on these attribution maps. We compare Attri-Net to five post-hoc explanation techniques and one inherently interpretable classifier on three chest X-ray datasets. We find that Attri-Net produces high-quality multi-label explanations consistent with clinical knowledge and has comparable classification performance to state-of-the-art classification models.
\end{abstract}

% \begin{keywords}
% Interpretable ML, Visual Feature Attribution, Multi-label Classification
% % List of keywords, comma separated.
% \end{keywords}

\section{Introduction}

%In recent years, deep neural networks have led to significantly improved performance on various medical image analysis tasks in experimental settings~\cite{litjens2017survey}. 
The clinical adoption of machine learning (ML) technology is hindered by the black-box nature of deep learning models. Their inscrutability may lead to a lack of trust~\cite{dietvorst2015algorithm}, or
%, perhaps more worryingly, 
blind trust among clinicians~\cite{tschandl2020human, gaube2021ai}, and may result in ethical as well as legal problems~\cite{grote2020ethics}. Therefore, transparency has been identified as one of the key properties for deploying machine learning technology in high-stakes application areas such as medicine~\cite{rudin2019stop}. 

%A variety of approaches have been proposed to better understand the decision mechanisms of machine learning models. 
The most commonly used category of techniques for understanding the decision mechanisms of ML models are \textit{post-hoc} methods which
%obtain a visual explanation of the prediction using an approximation of the decision process. 
apply a heuristic to a trained model trying to understand the decision mechanism retrospectively after the prediction is made.
Gradient-based techniques such as Guided Backpropagation~\cite{springenberg2014striving} perform local function approximation of the black-box model by differentiating the prediction with respect to the input pixels. 
The faithfulness of such methods to the decision mechanisms has recently been put into question by \citet{adebayo2018sanity} and \citet{arun2021assessing} who showed that explanations remain unchanged despite randomisation of network weights. Perturbation-based methods such as LIME \cite{ribeiro2016should}, or SHAP \cite{lundberg2017unified} also approximate the local decision function. These methods cannot currently produce explanations at the pixel level and are computationally demanding. 
% next sentence may have to go due to space limitations in the final version
%Moreover, these models may also not be faithful to the decision mechanism~\cite{white2019measurable}. 
Another line of work including Class Activation Mappings (CAM)~\cite{zhou2016learning} and GradCAM~\cite{selvaraju2017grad} attempts to construct neural network architectures from which the decision mechanism can be directly inferred. However, these techniques are limited by the spatial resolution of their explanations and do not explain the reasoning mechanism on a pixel-level. 
% Possibly reinclude this for camera ready
BagNet~\cite{brendel2019approximating} addresses this issue by severely restricting the global receptive field of the network which can negatively affect classification performance. %Additionally, explanations are partially generated by image patches that are not used for making the prediction. This means that, again, the explanations do not directly correspond to the decision mechanism. 
Placing attention modules at different depth throughout the network can also provide a measure of interpretability to individual feature maps~\cite{schlemper2019attention,yan2019melanoma}. A category of approaches highly related to our proposed method are counterfactual explanations which either try to answer the question ``What would the image look like if it belonged to a different class?'' ~\cite{schutte2021using,joshi2018xgems}, or exaggerate the features of the predicted class~\cite{cohen2021gifsplanation,singla2019explanation}. Other approaches in this category derrive classifications from an intermediate representation of the counterfactual generator~\cite{bass2020icam,cetin2022attri}.
% The following line will have to potentially go (but ideally we should cite the nemirovsky paper somehow
We also note that some techniques attempt to generate counterfactuals without the aim to explain a classifier~\cite{baumgartner2018vagan,nemirovsky2020countergan}.

\begin{figure}
    \centering
    \includegraphics[width=0.99\textwidth]{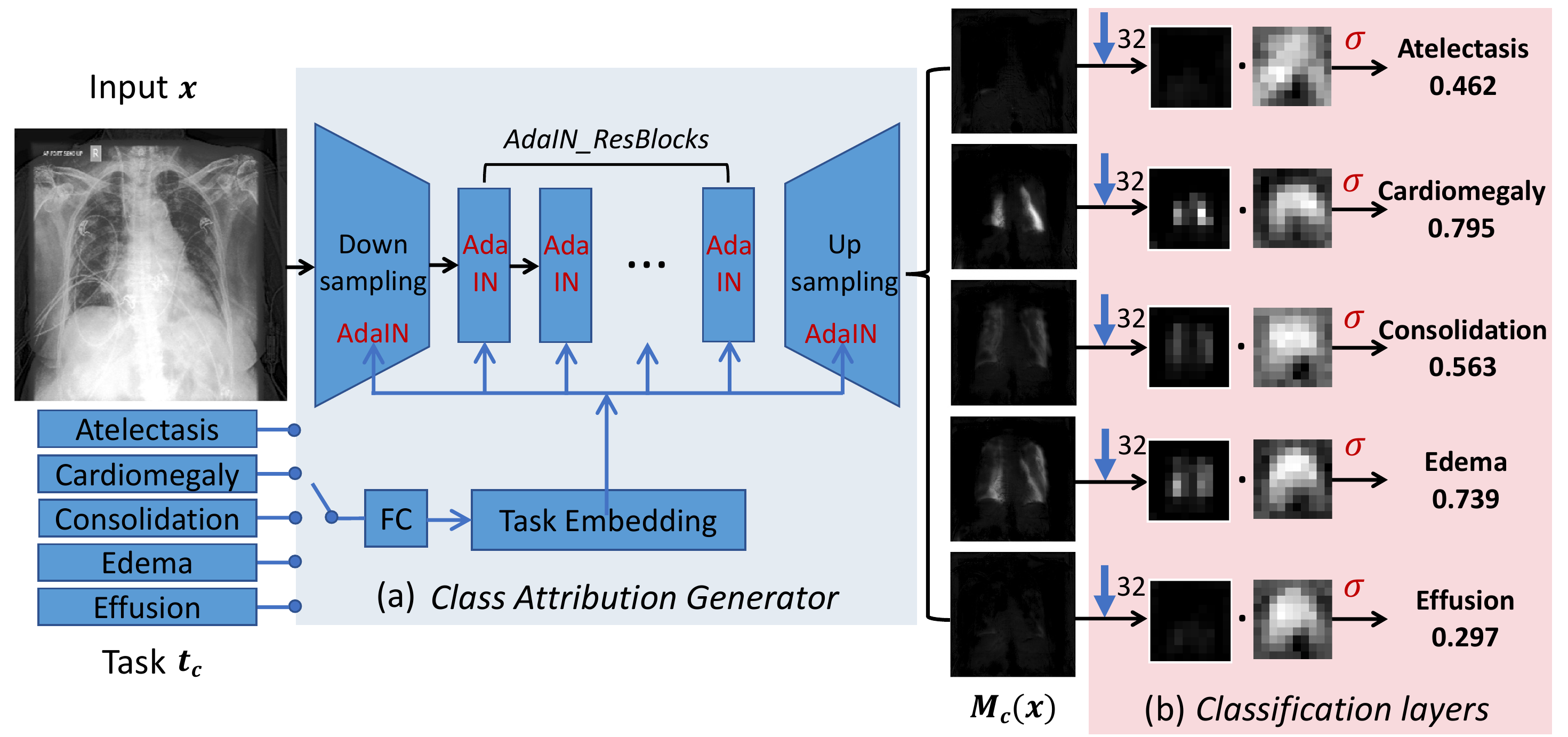}
    \vspace{-2mm}
    \caption{Attri-Net Framework. Given an input image and a task, our visual feature attribution generator (a) produces counterfactual attribution maps. Based on these maps a logistic regression classifier (b) produces the final prediction for each class.}
    \label{fig:framework}
    \vspace{-0.6cm}
\end{figure}

While \textit{post-hoc} explanations may appear reasonable there is no guarantee that they explain what the classifier actually does, and there is, in fact, growing evidence that they are not faithful to the actual decision mechanism~\cite{adebayo2018sanity,han2022explanation,white2019measurable}.
%It has recently been shown that no single linear post-hoc explanation method can perform optimally across all neighbourhoods~\cite{han2022explanation}. 
%In fact, if a post-hoc approximation was completely faithful to the model, it could simply replace the model~\cite{rudin2019stop}. 
In contrast, \textit{inherently interpretable} methods use prediction systems for which the decision mechanism is directly revealed to the user. These models are by definition faithful to the decision mechanism because the explanation \textit{is} the decision mechanism. Prior work includes methods in which the final predictions are directly based on human-interpretable concepts~\cite{alvarez2018towards,chen2020concept,koh2020concept}, prototypical representations of classes~\cite{chen2019looks,barnett2021interpretable}, or direct attribution to image patches~\cite{javed2022additive}. The recently proposed Convolutional Dynamic Alignment Networks (CoDA-Nets)~\cite{bohle2021convolutional} is, to our knowledge, the only existing model providing inherently interpretable visual explanations on the pixel-level. The method expresses network weights as a function of the input image in a way that allows them to formulate the networks decision for a specific input image as a linear classifier. We note that there are to our knowledge no inherently interpretable methods based on counterfactual explanations. 

The majority of visual explanation techniques were developed for binary or multi-class problems. Many clinical tasks, however, are multi-label problems, where multiple classes can apply simultaneously. For example, in chest X-ray diagnosis, which we study in this work, an image often contains multiple findings. In our experiments, we found that existing visual explanation techniques are not well-suited to this important type of decision problems. In particular, the explanations are not specific to the class and they tend to highlight similar regions for all classes, in some cases even when the class is not present in the image.

In this paper, we propose Attri-Net, an inherently interpretable visual explanation technique designed specifically for the multi-label scenario. Our model predicts class-specific counterfactual attribution maps as intermediate representations. The attribution maps, which are conceptually based on the visual feature attribution GAN (VA-GAN) approach introduced by \citet{baumgartner2018vagan}, represent residual images that contain all existing evidence of a class in an input image. Attri-Net then uses these class-specific attribution maps as input features in a final linear classification layer.
We evaluate Attri-Net on three widely used chest X-ray datasets and demonstrate that the method produces high-quality inherently interpretable explanations with a high class sensitivity while retaining classification performance comparable to state-of-the-art models. 

\section{Methods}
\label{sec:methods}
% TODO:
% - unfiy (residual) (counterfactual) (class) attribution maps. 
% - Is M the function or the generated map. Make sure it is precise. Should we introduce a generated map \mathbf{m}_c for clarity?

In this paper, we address the multi-label classification scenario with $C$ classes, where each class $c$ with label $y_c \in \{0,1\}$ can independently occur in an image, i.e. multiple co-existing medical findings are possible. In the following, we first introduce our method for generating counterfactual class attribution maps for each class $c$ (see Fig.\,\ref{fig:framework}a). Then, we show how a logistic regression classifier is used to obtain the final predictions based on those attribution maps (see Fig.\,\ref{fig:framework}b). Lastly, we explain how these two components are trained end-to-end in our proposed Attri-Net framework. 

\subsection{Counterfactual class attribution map generation} \label{method_diag}
\label{sec:methods-cams}
% adversarial loss - what makes it behave in that way?
The core of our method is an image-to-image network $M_c(\mathbf{x}): \mathbb{R}^{h \times w} \mapsto \mathbb{R}^{h \times w}$ which generates residual counterfactual class attribution maps for an input image $\mathbf{x}$. Intuitively, the output of $M_c$ represents how each pixel in the input should change in order to remove the effect of class $c$ from the image. Like \citet{baumgartner2018vagan}, we learn an additive mapping $M_c$ that makes the output image appear to come from the opposite class, that is
$$\hat{\mathbf{x}} = \mathbf{x} + M_c(\mathbf{x}) \text{\,,} $$
such that the generated counterfactual image $\hat{\mathbf{x}}$ is indistinguishable from images sampled from the distribution $p(\mathbf{x}|y_c=0)$ of real images \textit{not} containing class $c$. To ensure the correct behaviour of $M_c$, we simultaneously train a class-specific discriminator network $D_c$ to distinguish between real and fake images with $y_c=0$. Specifically, we use the Wasserstein GAN loss~\cite{arjovsky2017wasserstein,baumgartner2018vagan}.
%including the gradient penalty for improved training proposed by \citet{gulrajani2017improved}. 
Details on the optimisation of $D_c$ are given in Appendix~\ref{appendix:discriminator_training}. Given a discriminator function $D_c$ we can write the following adversarial loss term ensuring that $\hat{\mathbf{x}}$ is a realistic counterfactual not containing class $c$ and, by extension, that $M_c$ outputs realistic residual class attribution maps:
\begin{equation}
\mathcal{L}^{(c)}_{\text{adv}} = \E_{\mathbf{x}\sim p(\mathbf{x}|y_c=1)} [-D_c(\mathbf{x} + M_c(\mathbf{x})) ] \text{\,.}
\end{equation}
Some examples of generated counterfactuals $\hat{\mathbf{x}}$ are shown in Appendix~\ref{appendix:counterfactuals}.

% regularisation loss - sparsity
To discourage the network from attributing superfluous pixels not belonging to a given class, we additionally encourage the class attribution maps to be sparse using an $L_1$~regularization term similar to \citet{baumgartner2018vagan}. To further encourage the generator to produce smaller effects when the class is present in an image than when it is not present, we divide the loss into two differently weighted terms with a larger weight $\alpha_0$ for class-negative, and a smaller weight $\alpha_1$ class-positive examples, i.e.,  
\begin{equation}
\mathcal{L}^{(c)}_{\text{reg}} = \alpha_0 \E_{\mathbf{x}\sim p(\mathbf{x}|y_c=0)} [{\lVert M_c(\mathbf{x}) \rVert }_1] + \alpha_1 \E_{\mathbf{x}\sim p(\mathbf{x}|y_c=1)} [ {\lVert M_c(\mathbf{x}) \rVert }_1 ] \text{\,.}
\end{equation}
%where $\alpha_0, \alpha_1$ are hyperparameters chosen such that $\alpha_0 > \alpha_1$. 
%This ensures that images in which class $c$ is not present produce very little to no attribution.  
We use $\alpha_0 = 2, \alpha_1 = 1$ for all experiments in this paper. \vspace{1mm}

% Architecture
% - StarGAN (for both generator and discriminator)
The functions $M_c$, and $D_c$ are implemented as neural networks building on the StarGAN architecture~\cite{choi2018stargan} which produced superior results to alternative options we explored such as the original VA-GAN architecture. 
% - multi channel doesn't work --> task switching
Although it is feasible to design a network $M$ to produce class attribution maps for all labels as multiple output channels in a single forward pass, preliminary experiments revealed inadequate class attribution in the multi-label scenario.
% - AdaIN and changes to StarGAN. 
Instead, we build on the recently proposed task switching network~\cite{sun2021task} where adaptive instance norm (AdaIN) layers are used to switch between related tasks. In our work, tasks correspond to the generation of attribution maps for different classes. Each task is represented as a task vector $\mathbf{t}_c$ which is a one-hot encoding upsampled by a factor of 20 as in~\cite{sun2021task}. This encoding is then converted into a task embedding via a small fully connected network and fed to AdaIN layers which are placed throughout the network (as shown in Fig.\,\ref{fig:framework}a). The AdaIN layers then toggle the behaviour of the network. To combine this paradigm with the StarGAN architecture, we replaced all instance normalization layers of the original generator and discriminator networks with AdaIN layers. The architecture is described in greater detail in Appendix~\ref{appendix:architecture}. The mask generator and discriminator can now be expressed as $M_c(\mathbf{x}) = M(\mathbf{x}, \mathbf{t}_c)$, and $D_c(\mathbf{x}) = D(\mathbf{x}, \mathbf{t}_c)$, respectively. The class attribution maps for all labels can be obtained by repeated forward passes through $M$ while iterating through the $\mathbf{t}_c$ vectors of all classes. %Though, our method only takes advantage of the residual class attribution maps generated by $M$, we can also visualise the counterfactuals. Some examples are shown in Appendix~\ref{appendix:counterfactuals}.

\subsection{Classification using a logistic regression classifier}
\label{sec:methods-classification}

Given a class-specific counterfactual attribution map obtained using  $M(\mathbf{x}, \mathbf{t}_c)$, we want to predict the presence of class $c$ in an image. To achieve this, the respective attribution map is downsampled and used as input to a logistic regression classifier. That is, 
\begin{equation}
p(y_c|\mathbf{x}) = \sigma\Bigl(\sum_{i,j} w_{ij}^{(c)} \cdot S_\gamma (M(\mathbf{x}, \mathbf{t}_c))_{ij}\Bigr) \text{\,,}
\end{equation}
where $S_\gamma$ is a 2D average pooling operator that downsamples by a factor of $\gamma$, $w_{ij}^{(c)}$ denotes the weights associated to each pixel of the down-sampled attribution map for class $c$, and $\sigma$ is the sigmoid function. In preliminary experiments, we found $\gamma=32$ to perform robustly and we use this value for all experiments. 

The classifier is trained using a standard multi-label classification loss $\mathcal{L}^{(c)}_{\text{cls}}$, which is the sum of the binary cross entropy losses for each class. Note that, since our framework is trained end-to-end, $M$ also receives gradients from that loss and is thereby encouraged to create class attribution maps that are linearly classifiable. 

To further encourage the attribution maps to be discriminative for positive and negative examples of each class, we apply the center loss proposed by \citet{wen2016discriminative}, which has been shown to lead to more discriminative feature representations. Extending the idea, here, we define class centers $\mathbf{v}_{y_c=0}, \mathbf{v}_{y_c=1} \in \mathbb{R}^{h \times w}$ which are learnable and converge to prototypical representations of attribution maps corresponding to positive and negative instances of each class $c$. The center loss draws the class attribution maps closer to their respective class centers, resulting in a more clustered feature space where positive and negative samples are better linearly separable. The overall center loss can be written as
\begin{equation}
\mathcal{L}^{(c)}_{\text{ctr}} = \frac{1}{2} \left(
\E_{\mathbf{x}\sim p(\mathbf{x}|y_c=0)} \left[ {\lVert M(\mathbf{x}, \mathbf{t}_c) - \mathbf{v}_{y_c=0} \rVert }_2^2 \right] + 
\E_{\mathbf{x}\sim p(\mathbf{x}|y_c=1)} \left[ {\lVert M(\mathbf{x}, \mathbf{t}_c) - \mathbf{v}_{y_c=1} \rVert }_2^2 \right] \right)\text{\,.}
\end{equation}
The class center images are updated for each mini-batch in a separate gradient update interleaved with the updates of the network parameters as described by \citet{wen2016discriminative}. The final class center images as well as the logistic regression weights $w^{(c)}_{ij}$ may be used to further interpret the model's behaviour on a global level. Examples of both are shown in Appendix~\ref{appendix:global_interpretability}. However, we leave the exploration of global interpretability to future work. \vspace{1mm} 

\subsection{Training}
\label{sec:methods-training}

Our Attri-Net framework can be trained end-to-end with four loss terms enforcing our essential requirements: Firstly, the attribution map should preserve sufficient class relevant information such that a satisfactory classification result can be obtained. Secondly, the attribution maps should be human-interpretable.
The overall training objective for the class attribution generator $M$ with weight parameters $\varphi$ is given by

\begin{equation}
\min_\varphi \sum_c \lambda_{\text{cls}} \mathcal{L}_{\text{cls}}^{(c)} + \lambda_{\text{adv}} \mathcal{L}_{\text{adv}}^{(c)} + \lambda_{\text{reg}}\mathcal{L}_{\text{reg}}^{(c)} + \lambda_{\text{ctr}} \mathcal{L}_{\text{ctr}}^{(c)}  \text{\,,}
\end{equation}
where we use the hyperparameters $\lambda_*$ to balance the losses. We chose $ \lambda_{\text{cls}}=100, \ \lambda_{\text{adv}}=1, \ \lambda_{\text{reg}}=100, \ \lambda_{\text{ctr}}=0.01 $ for our experiments. An ablation study on the effect of the different losses can be found in Appendix~\ref{appendix:ablation}. During training, we repeatedly iterate through the different classes $c$ and, for each, sample two mini-batches, one containing positive examples of the current class and the other negative examples. Using these two batches we perform gradient updates for all losses before moving on to the next class. We use the ADAM optimizer~\cite{kingma2014adam} with a learning rate of $10^{-4}$ and a batch size of 4 to optimize our model. Furthermore, following \citet{wen2016discriminative}, we use stochastic gradient descent for updating the center loss parameters. Training converges within 72 hours on an Nvidia V100 GPU. %
After training we select the decision threshold which maximises the Youden-index (sensitivity + specificity - 1) for each class on the validation set. We also perform this step for the baseline methods.  

\begin{table}
    \centering
    \caption{Classification performance measured by area under the ROC curve (AUC).}
    \label{tab:auc}             
    \begin{tabular}{lccc}
        \hline
        \bfseries Model & \bfseries CheXpert & \bfseries ChestX-ray8 &\bfseries VindrCXR \\
        \hline
        ResNet50 \cite{azizi2021big} & 0.7687 & - & -\\
        SimCLR \cite{azizi2021big} & 0.7702 & - & -\\
        LSE \cite{ye2020weakly} & - & 0.7554 & -\\
        ChestNet \cite{ye2020weakly} & - & 0.7896 & -\\
  
        ResNet50 & 0.7727 & 0.7445 & 0.8986\\
        CoDA-Nets & 0.7659 & 0.7727 & 0.9322\\
        ours & 0.7405 & 0.7762 & 0.9405\\
  \hline
  \end{tabular}
  \vspace{-0.3cm}
\end{table}

\begin{table}
    \centering
    \caption{Comparison of class sensitivity scores.}
    \label{tab:class_sensitivity}             
    \begin{tabular}{lccc}
        \hline
        \bfseries Model & \bfseries CheXpert & \bfseries ChestX-ray8 &\bfseries VindrCXR \\
        \hline                  
        ResNet + GB        & 0.3183 & 0.3028 & 0.1727\\
        ResNet + GCam      & 0.1434 & 0.1570 &0.1931\\
        ResNet + LIME     & 0.2347 & 0.2609 &0.2422\\
        ResNet + SHAP    & 0.4745 & 0.4122 &0.3714\\
        %\revised{ResNet + Gifsplan.}    & \revised{0.2748} & \revised{0.5817} & \revised{0.4396}\\
        ResNet + Gifsplan.    & 0.2748 & 0.5817 & 0.4396\\
        CoDA-Nets   & 0.3576 & 0.4138 & 0.4464\\
        ours     & \textbf{0.4880} & \textbf{0.6160} & \textbf{0.5509}\\
        \hline
  \end{tabular}
  \vspace{-0.3cm}
\end{table}

\section{Experiments and Results} \label{sec:experiments}

% We evaluate our proposed Attri-Net in terms of classification performance, class sensitivity, as well as the qualitative human interpretability of explanations.
\noindent \textbf{Data.} We evaluated our proposed Attri-Net on the three widely used chest X-ray datasets CheXpert~\cite{irvin2019chexpert}, ChestX-ray8~\cite{wang2017chestx}, and VinDrCXR \cite{nguyen2020vindr}. Following \cite{irvin2019chexpert} and \cite{azizi2021big} for the CheXpert and ChestX-ray8 datasets we used the classes ``Atelectasis", ``Cardiomegaly", ``Consolidation", ``Edema", and ``Pleural Effusion". For the VinDr-CXR dataset, we selected the five pathologies with the highest number of samples, which were ``Aortic enlargement", ``Cardiomegaly", ``Pulmonary fibrosis", ``Pleural thickening", and ``Pleural effusion". We split all datasets into a training (80\%), testing (10\%) and validation (10\%) fold. Since the test set of Chexpert was not publicly available and the official validation set was small, we adopted the method used in \cite{azizi2021big} to split the train set into train, validation, and test sets. 

\vspace{1mm}
\noindent \textbf{Classification performance.} To assess the classification performance, we compared our model with the state-of-the-art inherently interpretable model CoDA-Nets \cite{bohle2021convolutional} as well as a standard black-box ResNet50 model. We also report the results of \citet{azizi2021big} and \citet{ye2020weakly} on CheXpert and ChestX-ray8, respectively. Attri-Net overall performed comparable to the state-of-the-art (see Tab.\,\ref{tab:auc}), with an area under the ROC curve that was slightly lower on CheXpert, similar to other methods on ChestX-ray8, and slightly better on VindrCXR. 

\begin{figure}
    \centering
    \includegraphics[width=0.95\textwidth]{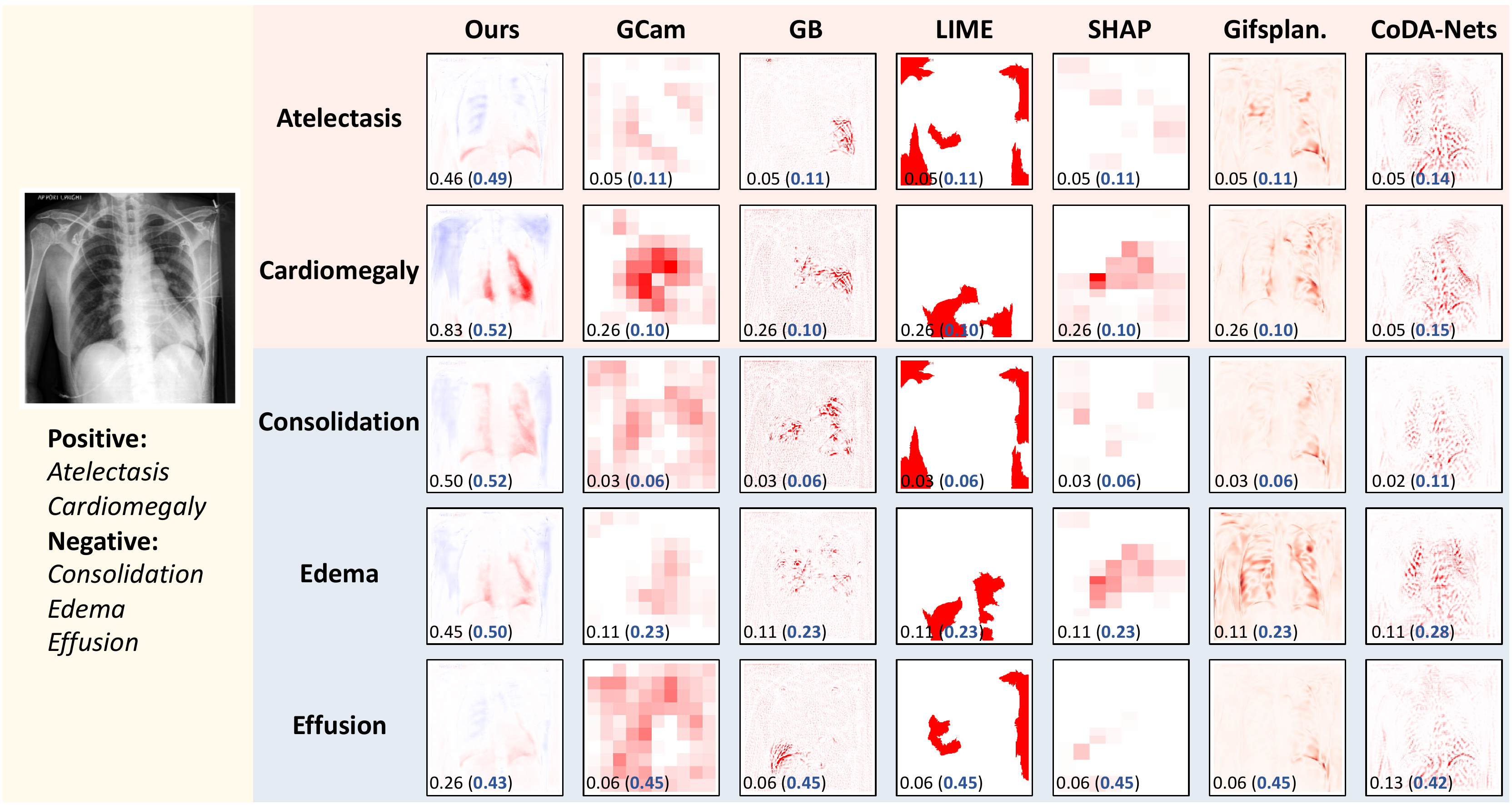}
    \vspace{-2mm}
    \caption{Visual comparison of explanations for an example image from the CheXpert dataset. Predicted class probabilities are indicated in the lower left corner of each attribution map with the respective decision threshold in parentheses.}
    \label{fig:class_sensitivity}
    \vspace{-0.6cm}
\end{figure}

\vspace{1mm}
\noindent \textbf{Interpretability.}
\citet{khakzar2021explaining} argue that if different areas of an image are responsible for predicting different classes, then also the explanations should be different. They coin this property ``class sensitivity''. 
% Similar to \citet{arun2021assessing} and \citet{krishna2022disagreement}, we observe that existing visual explanation techniques tend to provide highly similar explanations for all classes. This puts into question the usefulness of the explanation and is likely to reduce trust in the technology. 
In the context of multi-label classification, the explanation for an image containing a class should have higher attribution than an image where the class is absent. 
We measured class sensitivity following \citet{bohle2021convolutional} and created a series of $2 \times 2$ grids of explanations, where each grid contained only one positive example of a given class (see Appendix~\ref{appendix:sensitivity-grids} for example grids). We then represented class sensitivity by the sum of attributions in the positive example divided by the sum of all attributions in the grid.
%
%
% The class sensitivity score is calculated by creating a series of $2 \times 2$ image grids $\mathbf{g}^{(c)}\in \mathbb{R}^{2h \times 2w}$ consisting of the class effect maps for class $c$ of different input images. In each image grid, we include one class effect map for an input image containing class $c$, i.e. $\mathbf{m}^{(c_{pos})} = M(\mathbf{x}_{y_c=1},\mathbf{t}_c)$, and fill the other three spots with class attribution maps of images not containing class $c$, i.e. $\mathbf{m}^{(c_{neg})} = M(\mathbf{x}_{y_c=0}, \mathbf{t}_c)$. 
% The desired outcome is that only $\mathbf{m}^{(c_{pos)}}$ contains class attributions, and the three $\mathbf{m}^{(c_{neg})}$ show no, or fewer, attributions. 
% Note that we only constrain presence and absence of one class at a time. The input images may contain any of the other classes.
% Following \citet{bohle2021convolutional}, we define the class sensitivity score as 
% \begin{equation*}
% s^{(c)} = \sum\nolimits_{i,j} \mathbf{m}^{(c_{pos})}_{ij} / \sum\nolimits_{u,v}\mathbf{g}^{(c)}_{uv}.
% \end{equation*}
The optimal scenario where only the disease positive map contains any attributions, and disease negative attribution maps are blank, yields a sensitivity of $1$. We computed the average sensitivity over 200 grids for each class $c$. 

Our method led to a substantially and consistently higher class sensitivity than the inherently interpretable baseline, CoDA-Nets, across all datasets (see Tab.\,\ref{tab:class_sensitivity}). For the black-box ResNet, we compared five post-hoc explanations techniques, i.e. Guided Backpropagation~\cite{springenberg2014striving}, GradCAM~\cite{selvaraju2017grad}, LIME~\cite{ribeiro2016should}, SHAP~\cite{lundberg2017unified} and the recently proposed Gifsplanation~\cite{cohen2021gifsplanation}. The post-hoc methods varied considerably with SHAP and Gifsplanation performing comparably to CoDA-Nets, but substantially worse than our Attri-Net.

% We evaluate the class sensitivity for the inherently interpretable CoDA-Nets~\cite{bohle2021convolutional} as well as a ResNet with four different post-hoc explanation techniques. Those are Guided Backpropagation~\cite{springenberg2014striving}, GradCAM~\cite{selvaraju2017grad}, LIME~\cite{ribeiro2016should}, and SHAP~\cite{lundberg2017unified}. Our method led to a substantially and consistently higher class sensitivity than the other methods across all datasets (see Tab.\,\ref{tab:class_sensitivity}). 

\begin{figure}
    \centering
    \includegraphics[width=1.0\textwidth]{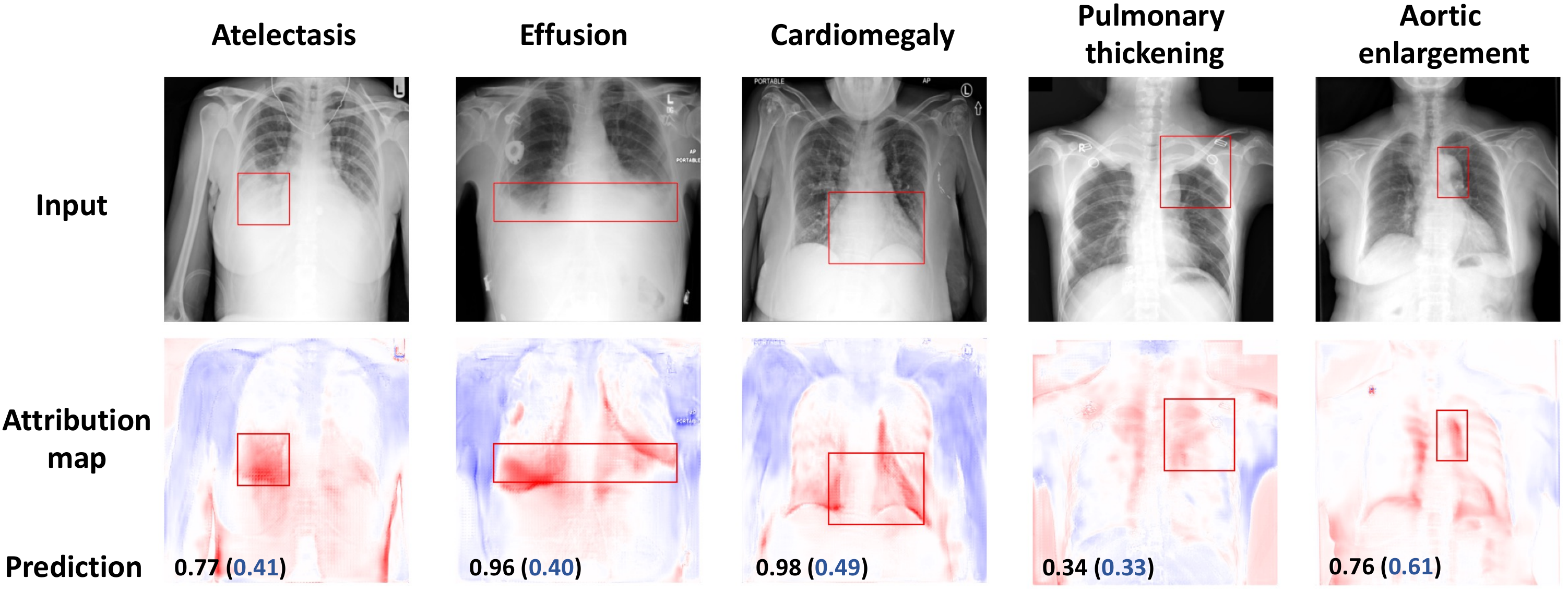}
    \vspace{-6mm}
    \caption{Attribution maps for samples from ChestX-ray8 and  VindrCXR with GT bounding boxes. Decision thresholds are given in parentheses.}
    \label{fig:bb_attri}
    \vspace{-0.3cm}
\end{figure}

Qualitative examination of example explanations supported these results. Our proposed Attri-Net produced class attribution maps that clearly highlight the parts of the underlying anatomy that support the respective classes (see Fig.\,\ref{fig:class_sensitivity} for a representative example from the CheXpert dataset). Moreover, the attributions for different classes were clearly distinct from each other, each one focusing on different anatomical areas. Examples from the ChestX-ray8 and VinDr-CXR datasets can be found in Appendix~\ref{appendix:visual_examples}.
In contrast, the inherently interpretable baseline, CoDA-Nets, produced visually similar attributions for all classes (rightmost column in Fig.\,\ref{fig:class_sensitivity}). We further observed that the baseline techniques were mostly not useful for identifying which parts of the anatomy contributed to a prediction. While Guided Backpropagation qualitatively provided the most useful explanations of the baselines, its attributions were very noisy as is typical for this technique.
We further examined Attri-Net explanations on example images of each class where pathology bounding boxes were available (Fig.\,\ref{fig:bb_attri}). Attri-Net generally highlighted regions associated with the respective pathologies, with particularly sensitive attribution maps when the final prediction was highly confident (i.e. the examples with atelectasis, effusion, and cardiomegaly). We also observed some relatively strong attributions in regions outside the bounding boxes. As our class attribution maps were based on counterfactuals that were designed to realistically remove all effects of a  pathology, we hypothesise they may have uncovered additional effects correlated with the classes which were not part of the clinical grading protocol. %This hypothesis could not be explored with the available bounding box annotations.

\section{Discussion and Conclusion}
%In this paper, w
We proposed Attri-Net, a novel inherently interpretable multi-label classifier and showed that it produces high-quality explanations substantially outperforming all baselines in terms of class sensitivity while retaining classification performance comparable to state-of-the-art black-box models. 
Explanations of the black-box model were highly dependent on the post-hoc technique, and fundamentally differed from each other even on the same image. This erodes trust in their capacity to provide necessary transparency in high-stakes applications and shows the need for inherently interpretable models such as ours, where the predictions are formed directly and linearly from visually interpretable class attribution maps.
%
% Since the final predictions are derived with a linear classifier which directly uses visually interpretable class attributions as features, the method as a whole is inherently interpretable. Thus, the explanations are by definition faithful to the model in contrast to post-hoc explanation models. 
%Post-hoc explanation methods, which constitutes the majority of related work, do not have this property and have recently been found and may in fact generate explanations unrelated to the model as recently demonstrated.

The qualitative and quantitative assessments in this paper suggest that our method provides useful explanations, but there remain important avenues for future work. 
% Remove this sentence again for final version and replace with the one below. 
%As pointed out by \citet{nemirovsky2020countergan} counterfactual explanations may sometimes be difficult to interpret. Therefore, the evaluation of different explanation types in human-ML collaboration setting will be a next crucial step. 
We believe a crucial step towards clinical impact is the evaluation of interpretable models in actual human-ML collaboration setting to test their usefulness with clinically relevant endpoints.

%
% Going forward we aim to evaluate our approach in different application areas and further explore the global interpretablity properties conferred by the center loss. 

\midlacknowledgments{Funded by the Deutsche Forschungsgemeinschaft (DFG, German Research Foundation) under Germany’s Excellence Strategy – EXC number 2064/1 – Project number 390727645. The authors acknowledge support of the Carl Zeiss Foundation in the project ``Certification and Foundations of Safe Machine Learning Systems in Healthcare" and the Hertie Foundation.
The authors thank the International Max Planck Research School for Intelligent Systems (IMPRS-IS) for supporting Susu Sun and Stefano Woerner. 
}

\bibliography{midl-samplebibliography}

\clearpage
\appendix

\section{Additional evaluations}

\subsection{Examples of counterfactual generations}
\label{appendix:counterfactuals}

\begin{figure}[h!]
    \centering
    \includegraphics[width=0.95\textwidth]{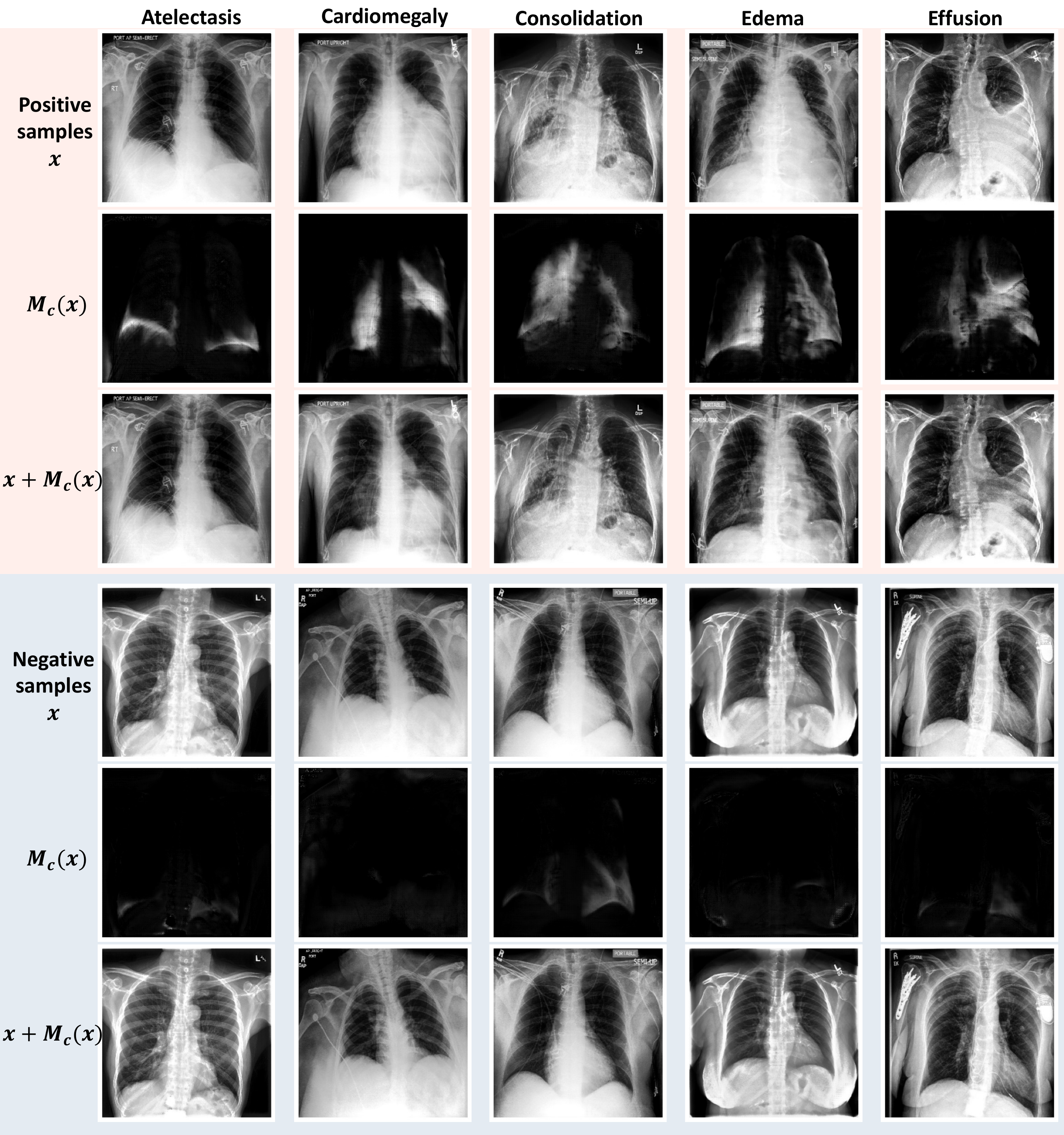}
    \vspace{-2mm}
    \caption{Counterfactual image generation. The examples in the top group of rows show input images containing evidence for different classes $c$. For those, the evidence for class $c$ is removed by adding $M_c(\mathbf{x})$. The bottom group of rows contains images with no evidence for class $c$. Those images remain mostly unchanged by adding the output of $M_c(\mathbf{x})$.}
    \label{fig:counterfactual_examples_chexpert}
    \vspace{-0.6cm}
\end{figure}

Examples of counterfactual images obtained by adding the class-specific visual attribution map to the input image, i.e. $\hat{\mathbf{x}} = \mathbf{x} + M_c(\mathbf{x})$, are shown in Fig.\,\ref{fig:counterfactual_examples_chexpert}. 

\subsection{Global interpretability} 
\label{appendix:global_interpretability}

A distinction is often made between local explanations, which explain the prediction for a specific input image, and \textit{global} explanations, which explain the decision mechanisms of the ML algorithm as a whole (i.e. for all input images). While the primary focus of our paper was on local interpretability, we may gain some global insights about the decision mechanism of the classifier through interpretation of the positive and negative class centers introduced in Section~\ref{sec:methods-classification}, as well as the class specific weights of the logistic regression classifier. The class centers capture some prototypical aspects of the respective classes, while the classifier weights can tell us which areas of the images the classifier is paying attention to for each class. Fig.\,\ref{fig: centers} shows the class centres for five diseases and the weights of the corresponding classifiers trained on the ChestX-ray8 dataset. 

\begin{figure}[htbp]
 % Caption and label go in the first argument and the figure contents
 % go in the second argument
\floatconts
  {fig: centers}
  {\caption{Attribution centers of disease and corresponding classifiers' weights provide a global explanation of Attri-Net.}}
  {\includegraphics[width=1.0
  \linewidth]{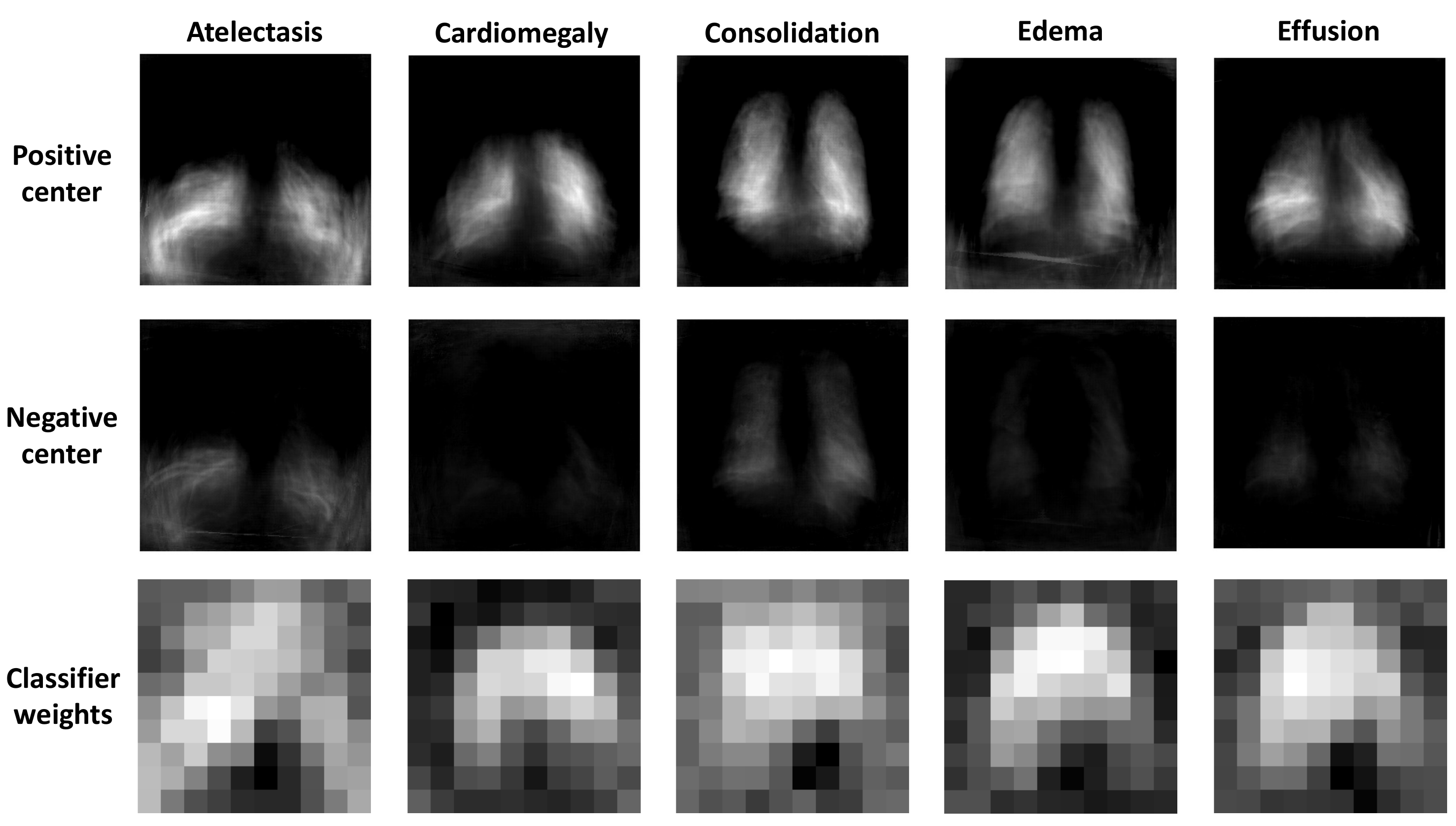}}
\end{figure}

\subsection{Ablation study of the loss terms}
\label{appendix:ablation}

An ablation study on the effects of the losses used for training Attri-Net can be found in Tab.\,\ref{tab:abla_models}. Example attributions for all combinations for an image from the ChestX-ray8 dataset are shown in Fig.\,\ref{fig: abla_vis}. 

\begin{table}[htbp]
\floatconts
  {tab:abla_models}%
  {\caption{Ablation study on the four losses.
  Evaluated on the Vindr-CXR dataset.}}%
  {\begin{tabular}{llcc}
  \hline
  \bfseries Model & \bfseries Loss terms &\bfseries Classification AUCs &\bfseries Class sensitivity\\
  \hline
  Attri-Net$_{\text{cls}}$ & $\mathcal{L}_{\text{cls}}$ &0.9339 &0.2516 \\
  Attri-Net$_{\text{cls\_adv}}$ & $\mathcal{L}_{\text{cls}}$ + $\mathcal{L}_{\text{adv}}$ & \textbf{0.9444} &0.1602\\
  Attri-Net$_{\text{cls\_adv\_reg}}$ & $\mathcal{L}_{\text{cls}}$ + $\mathcal{L}_{\text{adv}}$ + $\mathcal{L}_{\text{reg}}$ &0.9397 &0.5259 \\
  Attri-Net$_{\text{all}}$ & $\mathcal{L}_{\text{cls}}$ + $\mathcal{L}_{\text{adv}}$ + $\mathcal{L}_{\text{reg}}$+ $\mathcal{L}_{\text{ctr}}$ & 0.9405 & \textbf{0.5509}\\  
  \hline
  \end{tabular}}
\end{table}

\begin{figure}[htbp]
 % Caption and label go in the first argument and the figure contents
 % go in the second argument
\floatconts
  {fig: abla_vis}
  {\caption{Examples of attribution maps obtained for positive and negative samples of the disease cardiomegaly on ChestX-ray8 for different subsets of our losses.}}
  {\includegraphics[width=1.0\linewidth]{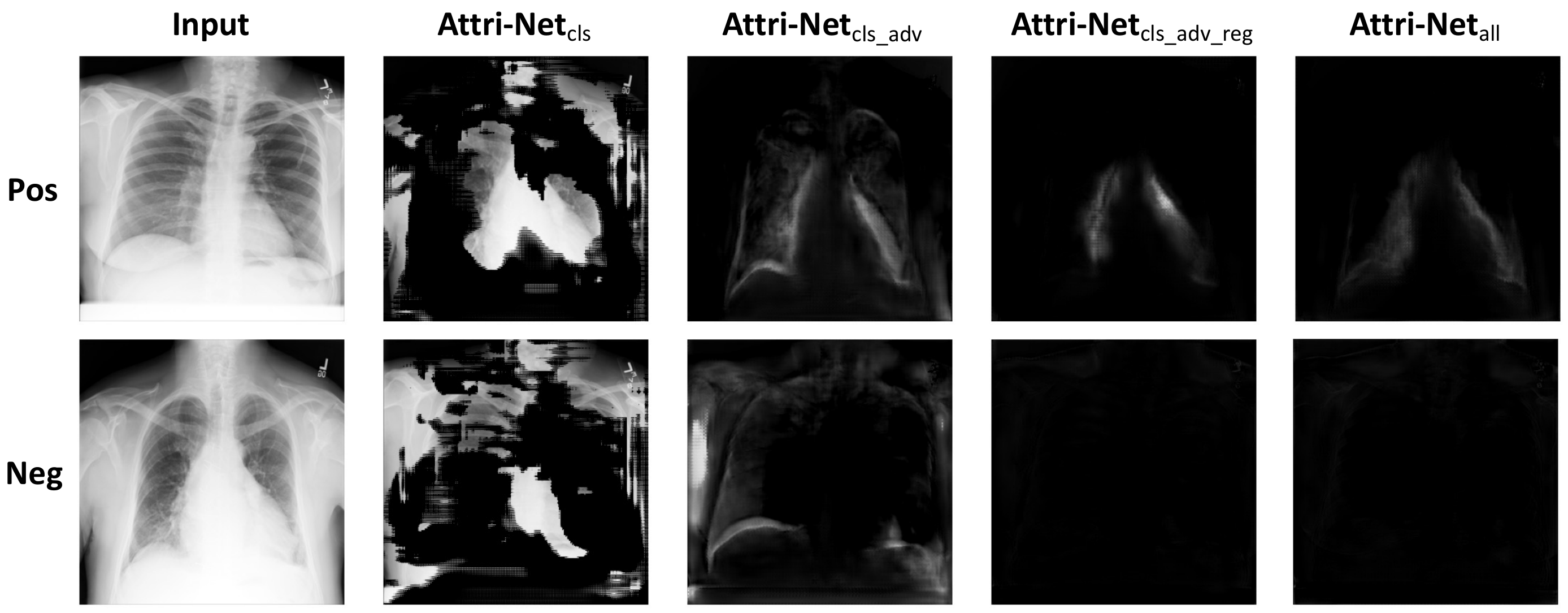}}
\end{figure}

\subsection{Class sensitivity image grids}
\label{appendix:sensitivity-grids}

% Some text referring to Fig.\,\ref{fig:sensitivity-grids-CoDA-net}. 
The class sensitivity evaluations in Section~\ref{sec:experiments} is based on class sensitivity grids as proposed by \citet{bohle2021convolutional}. In Fig.\,\ref{fig:sensitivity-grids-Attri-net}, Fig.\,\ref{fig:sensitivity-grids-CoDA-net}, and Fig.\,\ref{fig:sensitivity-grids-Gifsplanation} we show examples of such grids for all studied classes on the CheXpert dataset, for Attri-Net, CoDA-Nets and Gifsplanation, respectively. 

\begin{figure}[htbp]
 % Caption and label go in the first argument and the figure contents
 % go in the second argument
\floatconts
  {fig:sensitivity-grids-Attri-net}
  {\caption{Class sensitivity image grids obtained using Attri-Net. The first row shows image grids, the second row shows the respective attribution maps. P and  N denote class-positive and class-negative examples, respectively.}}
  {\includegraphics[width=1.0\linewidth]{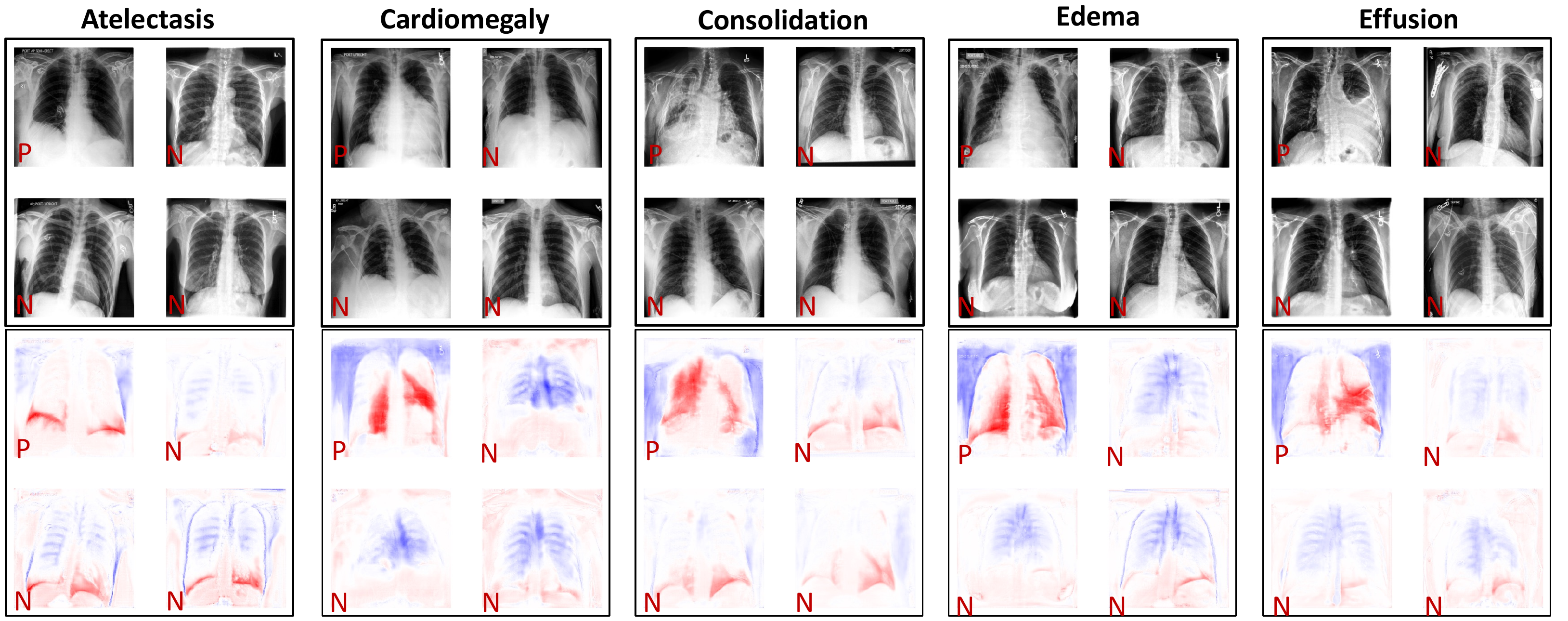}}
\end{figure}

\begin{figure}[htbp]
 % Caption and label go in the first argument and the figure contents
 % go in the second argument
\floatconts
  {fig:sensitivity-grids-CoDA-net}
  {\caption{Class sensitivity image grids obtained using CoDA-Nets. The first row shows image grids, the second row shows the respective attribution maps. P and  N denote class-positive and class-negative examples, respectively.}}
  {\includegraphics[width=1.0\linewidth]{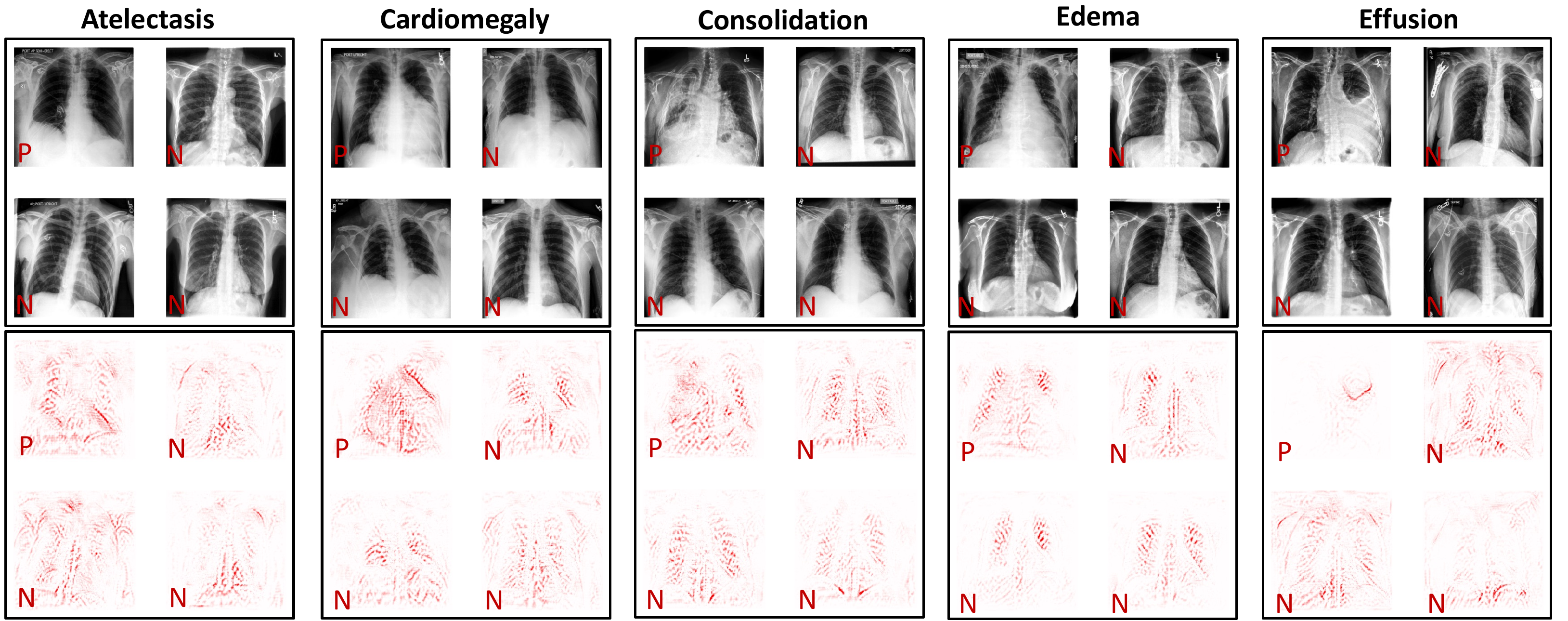}}
\end{figure}

\begin{figure}[htbp]
 % Caption and label go in the first argument and the figure contents
 % go in the second argument
\floatconts
  {fig:sensitivity-grids-Gifsplanation}
  {\caption{Class sensitivity image grids obtained using Gifsplanation. The first row shows image grids, the second row shows the respective attribution maps. P and  N denote class-positive and class-negative examples, respectively.}}
  {\includegraphics[width=1.0\linewidth]{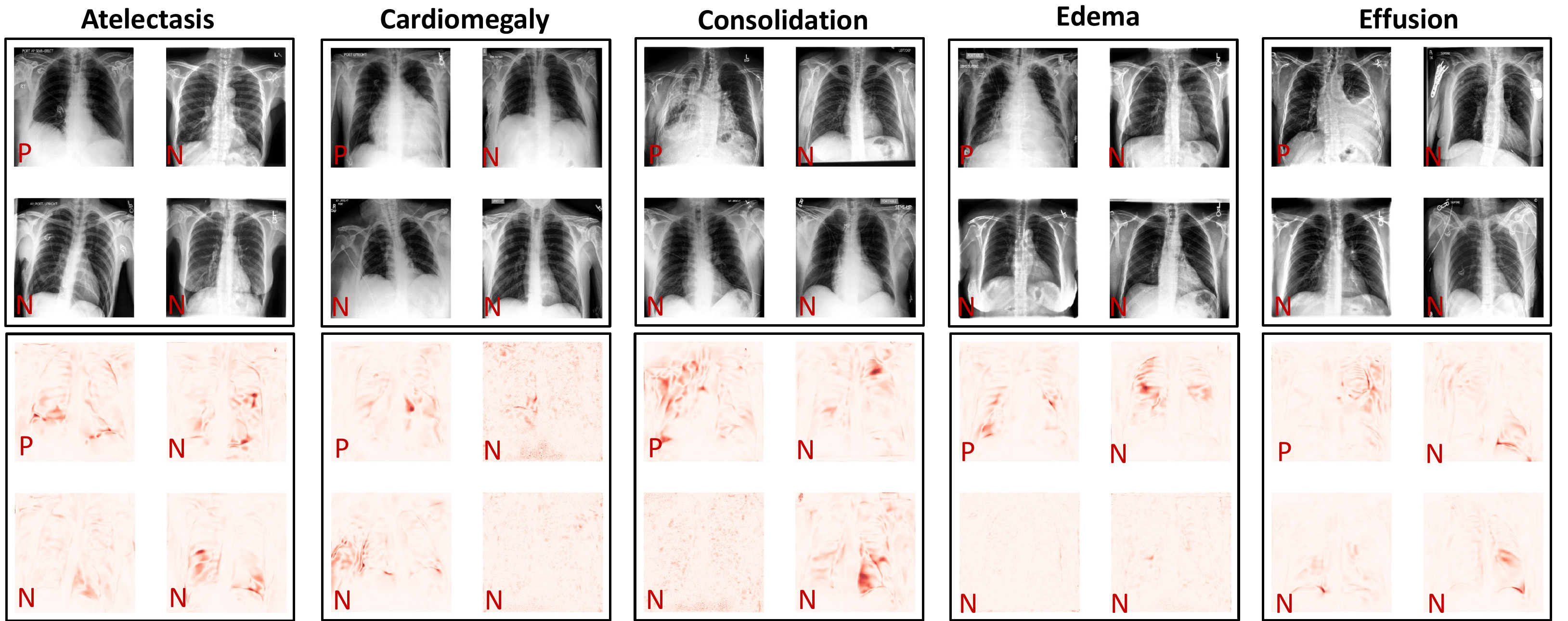}}
\end{figure}

\subsection{Example explanations for ChestX-ray8 and Vindr-CXR}
\label{appendix:visual_examples}

Fig.\,\ref{fig:sensitivity-nih} and Fig.\,\ref{fig:sensitivity-vindr} contain additional examples of visual attributions using all compared methods derived from the ChestX-ray8 and Vindr-CXR datasets, respectively.

\begin{figure}[htbp]
 % Caption and label go in the first argument and the figure contents
 % go in the second argument
\floatconts
  {fig:sensitivity-nih}
  {\caption{Explanations for an example image from the ChestX-ray8 dataset. }}
  {\includegraphics[width=1.0\linewidth]{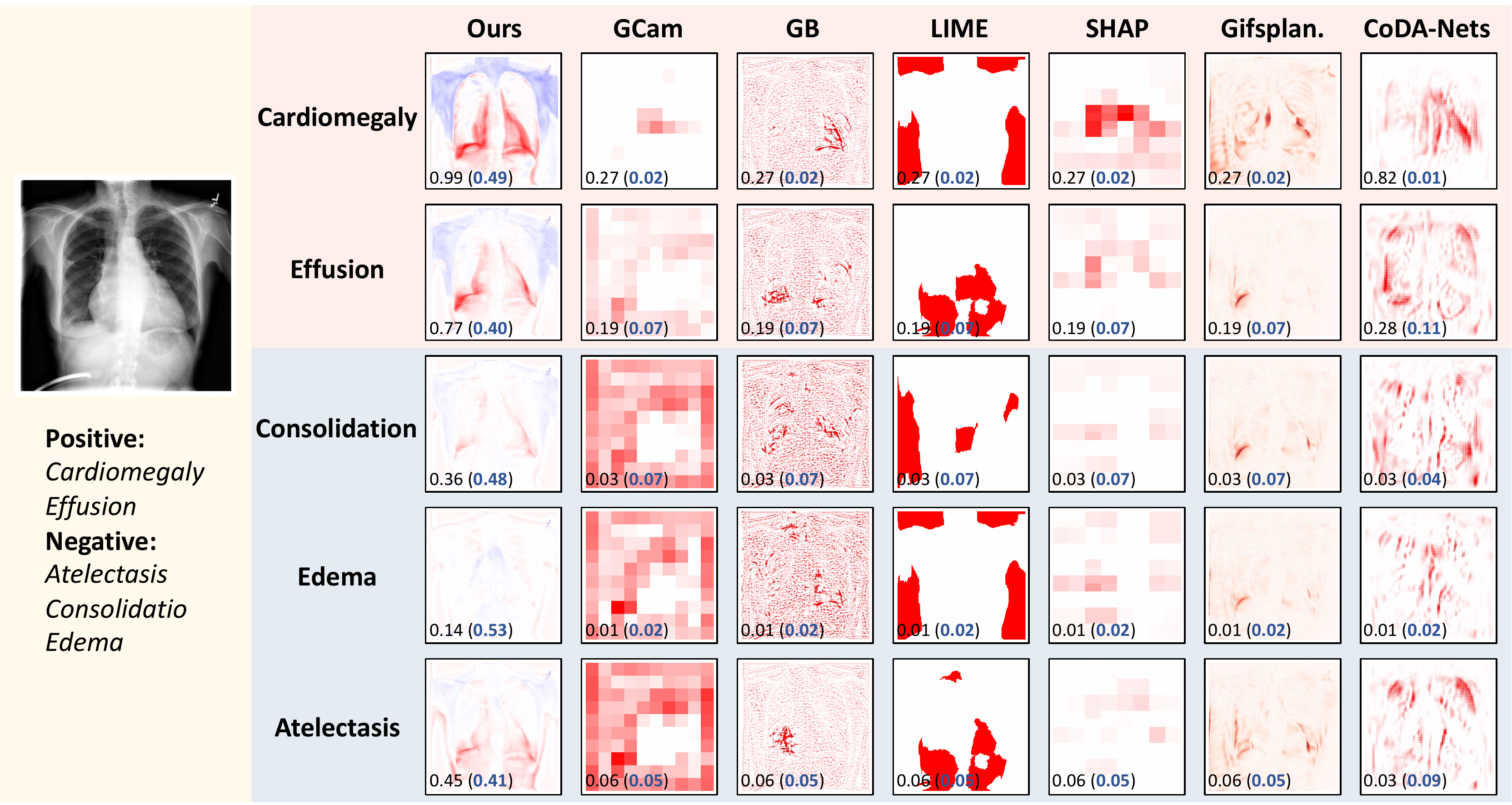}}
\end{figure}

\begin{figure}[htbp]
 % Caption and label go in the first argument and the figure contents
 % go in the second argument
\floatconts
  {fig:sensitivity-vindr}
  {\caption{Explanations for an example image from the VindrCXR dataset. }}
  {\includegraphics[width=1.0\linewidth]{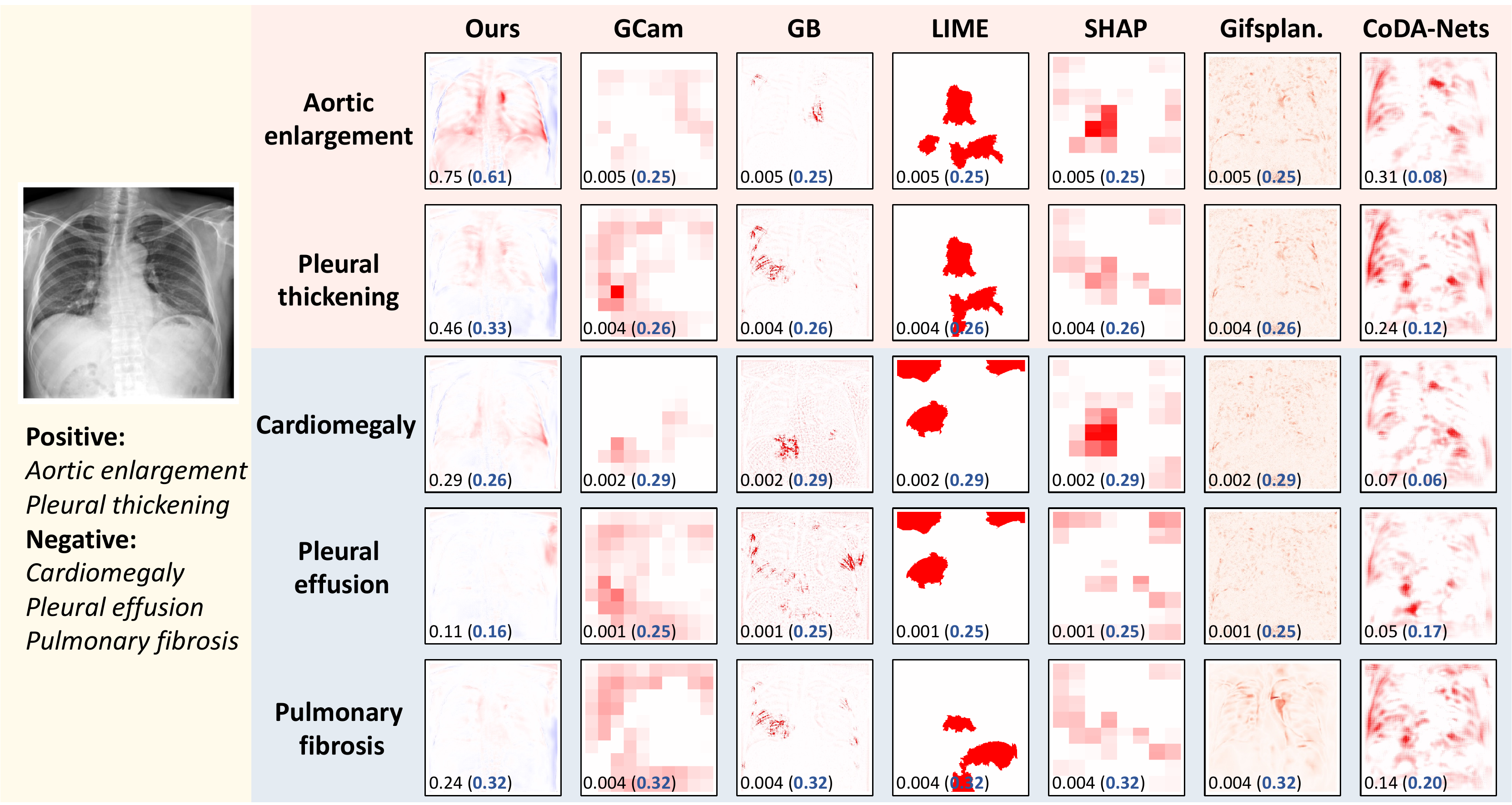}}
\end{figure}

\section{Additional training details}

\subsection{Discriminator training}
\label{appendix:discriminator_training}

The Attri-Net framework requires training a discriminator function $D$ in parallel to the class attribution generator $M$. The weight parameters $\theta$ of the discriminator are computed in separate gradient update steps using the Wasserstein GAN~\cite{arjovsky2017wasserstein} objective. The full discriminator optimisation objective is then given by
$$ \min_\theta \sum_c \E_{\mathbf{x} \sim p(\mathbf{x}|y_c=0)}[D_{c}(\mathbf{x}|\theta)] + \E_{\mathbf{x} \sim p(\mathbf{x}|y_c=1)}[D_{c}(\mathbf{x} + M_c(\mathbf{x})|\theta)] \text{\,,}$$
where we omitted the gradient penalty loss which ensures the discriminator fulfills the Lipschitz-1 constraint dictated by the Wasserstein GAN objective~\cite{gulrajani2017improved}.

\subsection{Network architecture}
\label{appendix:architecture}

The network architecture of the attribution map generator and the discriminator of the Attri-Net framework are shown in Tab.\,\ref{tab:arch_generator} and Tab.\,\ref{tab:arch_critic}, respectively. L refers to the length of input/output features, N are the number of output channels and K the kernel size.%For the class attribution generator and discriminator network, we use the adaptive instance normalization layer(AdaIN) to switch the networks to a specific task, and both networks use the activation function ReLU. There are some notations for the networks, 

\begin{table}[]
  \centering
  \caption{Attri-Net class attribution generator network architecture.}%
  \label{tab:arch_generator}
  \begin{adjustbox}{width=\textwidth}
\begin{tabular}{lcc}
\hline
\bfseries Layers     & \bfseries Input $\rightarrow$ Output       & \bfseries Layer information \\
\hline
\bfseries Task embedding layer & Task code $\mathbf{t}_c$ $\rightarrow$ Task embedding $\mathbf{t}_c^\prime$   & 8 $\times$ FC(L100,L100)\\
\hline 

                      &    & Ada\_Conv: CONV(N64, K7x7), AdaIN, ReLU\\
\bfseries Down-sampling & (Input image $\mathbf{x}$, $\mathbf{t}_c^\prime$) $\rightarrow$ $\mathbf{out}_{\text{down}}$       & Ada\_Conv: CONV(N128, K4x4), AdaIN, ReLU \\
                      &    & Ada\_Conv: CONV(N256, K4x4), AdaIN, ReLU \\
\hline
                      &    & Ada\_ResBlock: CONV(N256, K3x3), AdaIN, ReLU \\
                      &     & Ada\_ResBlock: CONV(N256, K3x3), AdaIN, ReLU \\
\bfseries Bottlenecks  & ($\mathbf{out}_{\text{down}}$ , $\mathbf{t}_c^\prime$) $\rightarrow$ $\mathbf{out}_{\text{bn}}$  & Ada\_ResBlock: CONV(N256, K3x3,), AdaIN, ReLU \\
                     &     & Ada\_ResBlock: CONV(N256, K3x3), AdaIN, ReLU \\
                     &     & Ada\_ResBlock: CONV(N256, K3x3), AdaIN, ReLU \\
                    &     & Ada\_ResBlock: CONV(N256, K3x3), AdaIN, ReLU \\
  \hline
                        & & Ada\_DECONV(N128, K4x4), AdaIN, ReLU \\
\bfseries Up-sampling & ($\mathbf{out}_{\text{bn}}$ , $\mathbf{t}_c^\prime$) $\rightarrow$ $\mathbf{out}_{\text{up}}$ & Ada\_DECONV(N64, K4x4), AdaIN, ReLU \\
          & & CONV(N1, K7x7) \\
  \hline
\bfseries Output layer & ($\mathbf{x}$, $\mathbf{out}_{\text{up}}$) $\rightarrow$  $M_c(\mathbf{x})$ &  $M_c(\mathbf{x}) = \text{tanh}(\mathbf{x}+\mathbf{out}_{\text{up}})-\mathbf{x}$\\
  \hline
    \end{tabular}
  \end{adjustbox}
\end{table}

\begin{table}[]
  \centering
  \caption{Attri-Net discriminator network architecture.}%
  \label{tab:arch_critic}
  \vspace{1mm}
  \begin{adjustbox}{width=\textwidth}
\begin{tabular}{lcc}
\hline
\bfseries Layers     & \bfseries Input $\rightarrow$ Output       & \bfseries Layer information \\
\hline
\bfseries Task embedding layer & Task code $\mathbf{t}_c$ $\rightarrow$ Task embedding $\mathbf{t}_c^\prime$   & 8 $\times$ FC(L100,L100)\\
\hline 
\bfseries Input layer & & Ada\_Conv: CONV(N64, K4x4), AdaIN, ReLU \\
& & Ada\_Conv: CONV(N128, K4x4), AdaIN, ReLU \\
& & Ada\_Conv: CONV(N256, K4x4), AdaIN, ReLU \\
\bfseries Hidden layers & ($\mathbf{x} \slash \hat{\mathbf{x}}$ , $\mathbf{t}_c^\prime$) $\rightarrow$ $\mathbf{out}_{\text{hid}}$  & Ada\_Conv: CONV(N512, K4x4), AdaIN, ReLU \\
& & Ada\_Conv: CONV(N1024, K4x4), AdaIN, ReLU \\
& & Ada\_Conv: CONV(N2048, K4x4), AdaIN, ReLU \\
  \hline
\bfseries Output layer  & $\mathbf{out}_{\text{hid}}$  $\rightarrow$ $\mathcal{L}^{(c)}_{\text{adv}}$ & CONV(N1, K3x3)\\
  \hline
    \end{tabular}
  \end{adjustbox}
\end{table}

\end{document}